\documentclass[11pt,a4paper]{article}
\usepackage[hyperref]{acl2019}
\usepackage{times}
\usepackage{latexsym}
\newcommand{\namecite}[1]{\newcite{#1}}
\usepackage{url}
\usepackage{multirow}

\aclfinalcopy

\title{Robust Machine Translation with 
Domain Sensitive Pseudo-Sources:
Baidu-OSU WMT19 MT Robustness Shared Task System Report}

\author{
  Renjie Zheng $^{* \hspace{0.5mm} \ddagger}$
  \qquad
  Hairong Liu \thanks{\quad Equal contribution} $^{\hspace{0.5mm} \dagger}$
  \qquad
  Mingbo Ma $^{\hspace{0.5mm} \dagger}$
  \qquad
  Baigong Zheng$^{\hspace{0.5mm} \dagger}$
  \qquad
  Liang Huang $^{\dagger,\ddagger}$
\\ 
$^{\dagger}$Baidu Research, Sunnyvale, CA\\
$^{\ddagger}$School of EECS, Oregon State University, Corvallis, OR \\
  {\tt \{zrenj11, lhrbss, cosmmb, zbgzbg2007, liang.huang.sh\}@gmail.com} \\
  }

\date{}

\begin{document}
\maketitle
\begin{abstract}
This paper describes the machine translation system
developed jointly by Baidu Research and Oregon State University 
for WMT 2019 Machine Translation Robustness
Shared Task.
Translation of social media is a very challenging problem,
since its style is very different from normal parallel
corpora (e.g. News)
and also include various types of noises. 
To make it worse, the amount of social media parallel corpora
 is extremely limited.
In this paper, we use a domain sensitive training method
which 
leverages a large amount of parallel data
from popular domains
together with a little amount of parallel data
from social media.
Furthermore, we generate a parallel dataset
with pseudo noisy source sentences
which are back-translated from monolingual data using
a model trained by a similar domain sensitive way.
We achieve more than 10 BLEU improvement in both
En-Fr and Fr-En translation compared with the baseline methods.
\end{abstract}

\section{Introduction}
Translation of social media is very challenging. First, there are various types of noises, such as abbreviations, spelling errors, obfuscated profanities, inconsistent capitalization, Internet slang and emojis  \cite{michel2018mtnt}.  Second, the amount of parallel data is limited. These characteristics of social media make existing neural machine translation systems extremely vulnerable.

The noise issue of social media has been investigated in some previous work \cite{baldwin2013noisy,eisenstein2013bad}.  Most recently, \citet{belinkov2017synthetic} demonstrated the vulnerability of neural machine translation system to both synthetic and natural noises. However, the noises tested in \cite{belinkov2017synthetic} are not real noises in social media. To our best knowledge, there seems to be a lack of translation methods systematically targeting noises in social media.

Existing neural machine translation systems are famous for their hungry of data. However, the amount of parallel data in social media domain is very limited. Just recently, a dataset collected from Reddit has been published and attracted a lot of attention \cite{michel2018mtnt}. The amount of data in this dataset is still very small, compared to the large amount of data from News domain. Naturally, how to utilize the large amount of parallel data from the News domain become a central problem in improving the translation of social meida.

In this paper, inspired by the success of back-translation technique \cite{sennrich2015improving}, we propose to learn a model to generate ``social-media-style'' translation in source language from clean sentences in target language. Since the amount of parallel data in social media domain is limited, we utilize the large amount of parallel data in News domain to help the training.  With this model, large mount of parallel data for back-translation can be generated from monolingual data in target language. In the final translation model, a special ``domain" symbol is added to indicate which domain the source sentence belonging to.

The contributions of this paper are multifold, and some important ones are highlighted below:
\begin{enumerate}
\item We found that ``social-media-style" sentences can be generated by training a translation model with different ``start-of-sentence" symbols for sentences in different domains in the decoder side. The model is trained with data from all domains,  especially News domain, which has a large amount of parallel data, but also adapted to the style in the domain of social media, even the amount of parallel data in social media is limited. As demonstrated by our experiments, generating ``social-media-style" sentences is crucial in the effectiveness of back-translation for training a translation model suitable for translating social media.
\item We illustrated that adding a domain symbol in source sentence improves the robustness of the model. This may be because the encoder learns some domain-specific features from input sentences.
\end{enumerate}

\section{Methods}

Noisy text translation is short of in-domain training data.
In this section, we present approaches to
leverage a large amount of out-of-domain
(e.g. News) dataset and monolingual data
paired with pseudo noisy source data from back-translation.

\subsection{Domain Sensitive Data Mixing}

To improve the translation model from limited parallel data,
we want to make the use of larger amount of out-of-domain data.
However, simply mixing the clean and noisy data will
make the whole training set unbalanced.
To differentiate the data from different domain,
we use different start symbol in source side.

The intuition of injecting domain label in source side is
based on the noise occurrence statistics from \cite{michel2018mtnt},
which shows much more spelling and grammar errors in the
source side of noisy text translation dataset.
Thus the clean and noisy start symbols work as a
meaningful sign of source text style for encoder.
Compared with the source side sentences, the human
translation of target side sentences are less noisier
with less spelling and grammar errors.

\subsection{Noisy Pseudo-Sources Generation with Back-Translation}

To further make the use of monolingual data, we
regard them as target data and generate it's corresponding
source data by back-translation \cite{sennrich2015improving}.
However, different from \namecite{sennrich2015improving} who uses
this method in both clean source and target sentences, 
the source side sentences in our test set is much noisier 
than target side (as mentioned in previous subsection).
Therefore, we reverse the source and target sentences
where the noisy source sentences becomes target and
cleaner target sentences becomes source.
For example, to generate noisy pseudo French source sentences
for English monolingual data, we train a En-Fr translation
model which takes the noisy French source sentences in
Fr-En noisy dataset as target, and the corresponding
paralleled English target sentences as source.
In this way, the model will learned how to inject
noise into the target side.
Similar to previous domain sensitive method, we
include out-of-domain clean data during the
training of this noisy translation model
and differentiate them by different start symbol
int target side.

\subsection{Ensemble}

In our experiments with relatively small training dataset,
the translation qualities of models with different initializations
can vary notably.
To make the performance much more stable and improve the translation quality,
we ensemble different models during decoding to achieve better translation.

To ensemble, we take the average of all model outputs:
\begin{equation}
\hat{y_t} = \sum_{i = 1}^N {\hat{y^i_t} \over N}
\end{equation}
where $\hat{y^i_t}$ denotes the output distribution of $i$th model
at position $t$.
Similar to \namecite{zhou2017neural} and \namecite{zheng2018ensemble}, we can ensemble models
trained with different architectures and training algorithms.

\section{Experiments}

To investigate the empirical performances of our proposed methods,
we conduct experiments on MTNT dataset \cite{michel2018mtnt}
using Transformer \cite{vaswani+:2017}.

We first apply BPE~\cite{sennrich+:2015} on both sides in order 
to reduce the vocabulary for both source and target sides.
We then exclude the sentences pairs whose length are longer than 256 words or subwords.
We use length reward \cite{huang2017finish} to find the optimal
target length.

Our implementation is adapted from PyTorch-based OpenNMT \cite{klein+:2017}.
Our Transformer's parameters are as the same as the base model's parameter settings in 
the original paper \cite{vaswani+:2017}.

In all experiments, our evaluation uses sacreBLEU
\footnote{https://github.com/mjpost/sacreBLEU},
a standardized BLEU score evaluation tool  by \namecite{post2018call}.
We specify the $\tt intl$ tokenization option during BLEU evaluation.
We also uses detokenization and normalization tools in Moses.

\begin{table}[!]\centering
\begin{tabular}{|l|c|c|c|c|}\hline
   & Training & Validation & Test \\\hline
Clean &  2,207,962 &       -    &   -    \\\hline
Monolingual & 26,485 & -    &   -    \\\hline
Noisy &   36,058     &     852      &   1,020    \\\hline
\end{tabular}
\caption{Statistics of En2Fr Dataset. Monolingual data is French only.}
\label{tab:en2fr_dataset}
\end{table}

\begin{table}[!]\centering
\begin{tabular}{|l|c|c|c|c|}\hline
      &  Training & Validation & Test \\\hline
Clean &  2,207,962  &      -    &   -    \\\hline
Monolingual & 2,244,020 & -    &  -     \\\hline
Noisy &     19,161  &   886        &    1,022   \\\hline
\end{tabular}
\caption{Statistics of Fr2En Dataset, Monolingual data is English only.}
\label{tab:fr2en_dataset}
\end{table}

Table \ref{tab:en2fr_dataset} and \ref{tab:fr2en_dataset} show
statistics of En2Fr and Fr2En datasets.
For both En-Fr and Fr-En dataset, the clean parallel data is from WMT15
news translation task.
The noisy data is from \cite{michel2018mtnt} collected from
social network. 
Except the French and English monolingual data from WMT15 news
translation task, we also make the use of English portion of
parallel data from KFTT, TED and JESC used in \cite{michel2018mtnt}.

\subsection{Noisy Data Generation}

\begin{table}[!]\centering
\begin{tabular}{|l|c|c|c|c|}\hline
	      &   En2Fr & Fr2En \\\hline
Domain Insensitive & 31.3  &  34.6  \\\hline
Domain Sensitive   & 35.7  &  39.5 \\\hline
\end{tabular}
\caption{Results of noisy data generation. We reverse 
the source and target direction of MTNT
Fr2En (En2Fr) dev-set to evaluate 
the ability of noisy data generation for En2Fr (Fr2En).}
\label{tab:noisy}
\end{table}

To make use of monolingual target data, we want to generate
the corresponding parallel pseudo noisy source data 
and put them into training set.
Table \ref{tab:noisy} shows the performance of our noisy
data generation models.
In this experiment, we mix the clean and noisy dataset as the 
training set,
but use the target sentences in reversed direction
of noisy dataset (training, validation, test) set
as source and source sentences as target.
The domain insensitive method simply mix the clean and noisy dataset
in training while the domain sensitive method
differentiate the clean and noisy dataset in target side
by starting with different symbol (e.g. $\tt<clean\_s>$, $\tt<noisy\_s>$).
The experiment shows that the domain sensitive
method can outperform the domain insensitive method with a large margin.

\subsection{Methods Comparison}

\begin{table*}[!]\centering
\begin{tabular}{|l|l|c|c|c|}\hline
	      &  Methods      & En-Fr & Fr-En \\\hline
\multirow{2}{*}{\shortstack[l]{Baseline}} & MTNT $\dagger$  & 21.8  & 23.3 \\
 & + tuning $\dagger$ & 29.7  & 30.3   \\\hline
\multirow{2}{*}{\shortstack[l]{Domain \\ Insensitive}}    & Mix training  &33.4  &34.5   \\
 & + Back translation  &33.7  &34.3   \\\hline
\multirow{3}{*}{\shortstack[l]{ Domain \\ Sensitive}}   & Mix training  & 36.3  & 38.7  \\
 & + Noisy back translation  & 38.4 & 41.0 \\
          & + Ensemble      & 40.4 & 42.3  \\\hline
\end{tabular}
\caption{Results of different methods on test-set. $^{\dagger}$\cite{michel2018mtnt}}
\label{tab:test}
\end{table*}

Table \ref{tab:test} shows the final results of different methods on test set.
Similar with the previous experiments, the domain
insensitive methods simply mix all the clean, noisy training data.
The performance has a little improvement in En-Fr by
adding the monolingual data paired with the pseudo source data
generated by the model trained in previous experiments.
To differentiate the clean and noisy dataset, we
assign different label at the start of them and the performance is thus boosted about 3 to 4 BLEU score.
We further generate pseudo noisy source data from the monolingual 
target with the model using the domain sensitive method in previous 
experiment.
By adding these noisy back-translation data, we achieve
more than 2 BLEU improvement.
Our final submission ensembles 5 models trained with the domain
sensitive method and including the noisy back translation data.

\subsection{Final Results}

\begin{table*}[!tbh]\centering
\begin{tabular}{|l|c|c|c|c|}\hline
	      & BLEU & BLEU-cased & BEER & CharacTER \\\hline
NLE       & 48.8 &  47.9      & 0.676    &   0.364  \\\hline
CUNI      & 45.8 &  44.8      & 0.654    &  0.395  \\\hline
BD-OSU$^*$ &44.5& 43.6      & 0.641    &  0.499  \\\hline
JHU       &  41.2&  40.2      &  0.624   &  -      \\\hline
CMU       &  32.8&  32.2      &  0.573   &  0.514  \\\hline
FOKUS$^\dag$   &  30.8&  29.9      &  0.530   &  0.574  \\\hline
MTNT & 26.2&25.6      &  0.529   &  0.550  \\\hline
IITP-MT   &  25.5&  20.8      &  0.499   &  0.594  \\\hline
\end{tabular}
\caption{Semi-blind test results of Fr-En.  $^*$Our submission. $^\dag$Unconstrained.}
\label{tab:fr-en_test}
\end{table*}

\begin{table*}[!tbh]\centering
\begin{tabular}{|l|l|c|c|c|}\hline
	      & BLEU & BLEU-cased & BEER  & CharacTER \\\hline
NLE       & 42.0 &  41.4      & 0.626    &   0.446  \\\hline
CUNI      & 39.1 &  38.5      & 0.605    &  0.483  \\\hline
BD-OSU$^*$  &  37.0 & 36.4      & 0.599    &  0.512  \\\hline
FOKUS$^\dag$   &  24.8 &  24.2     &  0.515   &  0.619  \\\hline
MTNT      & 22.5 & 22.1      &  0.498   &  0.621  \\\hline
CMU      & 20.8 & 20.4      &  0.488   &  0.622  \\\hline
IITP-MT   &  20.7&  19.2      &  0.492   &  0.619  \\\hline
SFU   &  19.4&  19.1      &  0.491   &  0.614  \\\hline
\end{tabular}
\caption{Semi-blind test results of En-Fr. $^*$Our submission. $^\dag$Unconstrained.}
\label{tab:en-fr_test}
\end{table*}

\begin{table*}[!tbh]\centering
\begin{tabular}{|l|l|c|c|c|}\hline
	        & En-Fr & En-Fr & Fr-En  & Fr-En \\\hline
	        & Score & Rank  & BLEU   & Rank \\\hline
BD-OSU$^*$ & 71.5  & 2     & 80.6   & 3  \\\hline
CMU         & -     & -     & 58.2   & 6  \\\hline
CUNI        & 66.3  & 3     & 82.0   & 2  \\\hline
JHU         & -     & -     & 76.3   & 4  \\\hline
NaverLabs   & 75.5  & 1     & 85.3   & 1  \\\hline
FOKUS$^\dag$& 52.5  & 4     & 62.6   & 5  \\\hline

\end{tabular}
\caption{Human judgments over all submitted systems (the higher the better) $^*$Our submission. $^\dag$Unconstrained.}
\label{tab:human}
\end{table*}

Table \ref{tab:fr-en_test} and Table \ref{tab:en-fr_test}
show the final results of our submission in Fr-En and En-Fr.
Our system ranks third in both directions.
Table \ref{tab:human} shows the human judgments over all submitted systems
which are done by \namecite{li2019findings} who also analyze and discuss all
submitted systems.


\section{Related Work}
The method proposed in this paper is a kind of domain adaptation technique. There are many previous work on domain adaptation for machine translation \cite{britz2017effective, wang2017instance, chu2017empirical, chu2018survey},  which leverages out-of-domain parallel corpora and in-domain monolingual corpora to improve translation. The difference between our method and previous work lies in that we use back-translation \cite{sennrich2015improving} for domain adaptation.
Different from some previous work using
 adversarial training \cite{liu2017adversarial}
or different attention \cite{zheng2018same} to
 differentiate multiple tasks,
we simply assign different starting symbol for multiple
tasks \cite{lample2018phrase}.

A similar method was proposed in \cite{xie2018noising} in the context of grammar correction, where a model is trained to add noises on original sentences to produce noisy sentences. However, instead of learn how to generate arbitrary ``noises", our goal is to learn ``social-media-style" translations.
\namecite{singh2019improving} injects artificial noise in the clean data
according to the distribution of noisy data.
\namecite{liu2018robust} propose to leverage phonetic information
to reduce the noises in data.

Another group of work related to this paper is data augmentation
in machine translation.
Although data augmentation is very popular in general learning
tasks, such as image processing, it is non-trivial to 
do so in machine translation because even slight modifications 
of sentences can make huge difference in semantics.
To our best knowledge, there are two categories of successful 
data augmentation approaches for machine translation.
The first one is based on back-translation (\cite{sennrich2015improving})
which augments monolingual data into training set.
The second one is based on word replacement, such as \cite{sennrich2016edinburgh} and \cite{wang2018switchout}.
\namecite{zheng2018multi} make the use of multiple references
and generates even more pseudo-references and achieve improvement in both
machine translation and image captioning.

\section{Conclusions and Future Work}
In this paper, we proposed a method to improve the translation of social media. The style of social media is very unique, and is very different from the style of widely researched News sentences.
The core part of our method is to generate useful parallel data for back-translation, that is, generating synthetic in-domain parallel data. To achieve this goal, we proposed a method to generate  ``social-media-style" source sentences from monolingual target sentences. We also distinguish the domain of source sentences by inserting a domain symbol into source sentences. Both techniques are proven to be extremely useful in the scenario of translating social media. Finally, we utilized the ensemble to  further boosts the translation performance.

The noises in social media are mostly introduced by 
human mistakes.
There are some other cases that noises in source side are
introduced by systems, such as ASR in speech-to-text
translation \cite{liu2019end}.
We plan to further investigate this domain sensitive method
on these tasks,
even on speech-to-text simultaneous translation
\cite{ma2018stacl, zheng2019simultaneous}.


\bibliography{robustcomp}
\bibliographystyle{acl_natbib}

\end{document}